\documentclass[10pt,twocolumn,letterpaper]{article}

 \usepackage[pagenumbers]{cvpr} 

\usepackage[dvipsnames]{xcolor}

\definecolor{cvprblue}{rgb}{0.21,0.49,0.74}
\usepackage[pagebackref,breaklinks,colorlinks,citecolor=cvprblue]{hyperref}
\usepackage{comment}
\usepackage{multirow}
\usepackage{algorithm}
\usepackage{algorithmic}
\usepackage{mathtools}
\newcounter{thmcount}
\usepackage{comment}

\newtheorem{definition}[thmcount]{Definition}

\usepackage{csquotes}
\def\paperID{*****} 
\def\confName{CVPR}
\def\confYear{2024}

\title{Wasserstein Distance-based Expansion of Low-Density Latent Regions for Unknown Class Detection}

\newtheorem{remark}{Remark}[section]
\author{Prakash Mallick, Feras Dayoub, Jamie Sherrah\\
Australian Institute of Machine Learning\\
North Terrace, Adelaide SA 5000\\
{\tt\small \{prakash.mallick, feras.dayoub, jamie.sherrah\}@adelaide.edu.au}
}

\begin{document}
\maketitle
\begin{abstract}
\textcolor{black}{This paper addresses the significant challenge in open-set object detection (OSOD): the tendency of state-of-the-art detectors to erroneously classify unknown objects as known categories with high confidence. We present a novel approach that effectively identifies unknown objects by distinguishing between high and low-density regions in latent space. Our method builds upon the Open-Det (OD) framework, introducing two new elements to the loss function. These elements enhance the known embedding space's clustering and expand the unknown space's low-density regions. The first addition is the Class Wasserstein Anchor (CWA), a new function that refines the classification boundaries. The second is a spectral normalisation step, improving the robustness of the model. Together, these augmentations to the existing Contrastive Feature Learner (CFL) and Unknown Probability Learner (UPL) loss functions significantly improve OSOD performance. Our proposed OpenDet-CWA (OD-CWA) method demonstrates: a) a reduction in open-set errors by approximately $17\%-22\%$, b) an enhancement in novelty detection capability by $1.5\%-16\%$, and c) a decrease in the wilderness index by $2\%-20\%$ across various open-set scenarios. These results represent a substantial advancement in the field, showcasing the potential of our approach in managing the complexities of open-set object detection. Code is available at GitHub repo \href{https://github.com/proxymallick/OpenDet_CWA}{\textit{https://github.com/proxymallick/OpenDet\_CWA}}.
}
\end{abstract}

\section{Introduction}
\label{sec:intro}
Object detectors typically operate under a closed-set assumption, expecting that testing scenarios will only involve object categories from the training datasets. However, in real-world applications, these systems often encounter unseen object categories, leading to significant performance degradation. This limitation becomes evident in scenarios where detectors misclassify unfamiliar objects with high confidence~\cite{scheirer2013}. Addressing this, Open-set Object Detection (OSOD)~\cite{Dimity_dropout,OpenWorld,han2022opendet} has emerged, focusing on detecting both known and unknown objects in diverse conditions.

\begin{figure}[t]%
     {\includegraphics[width=8.6cm]{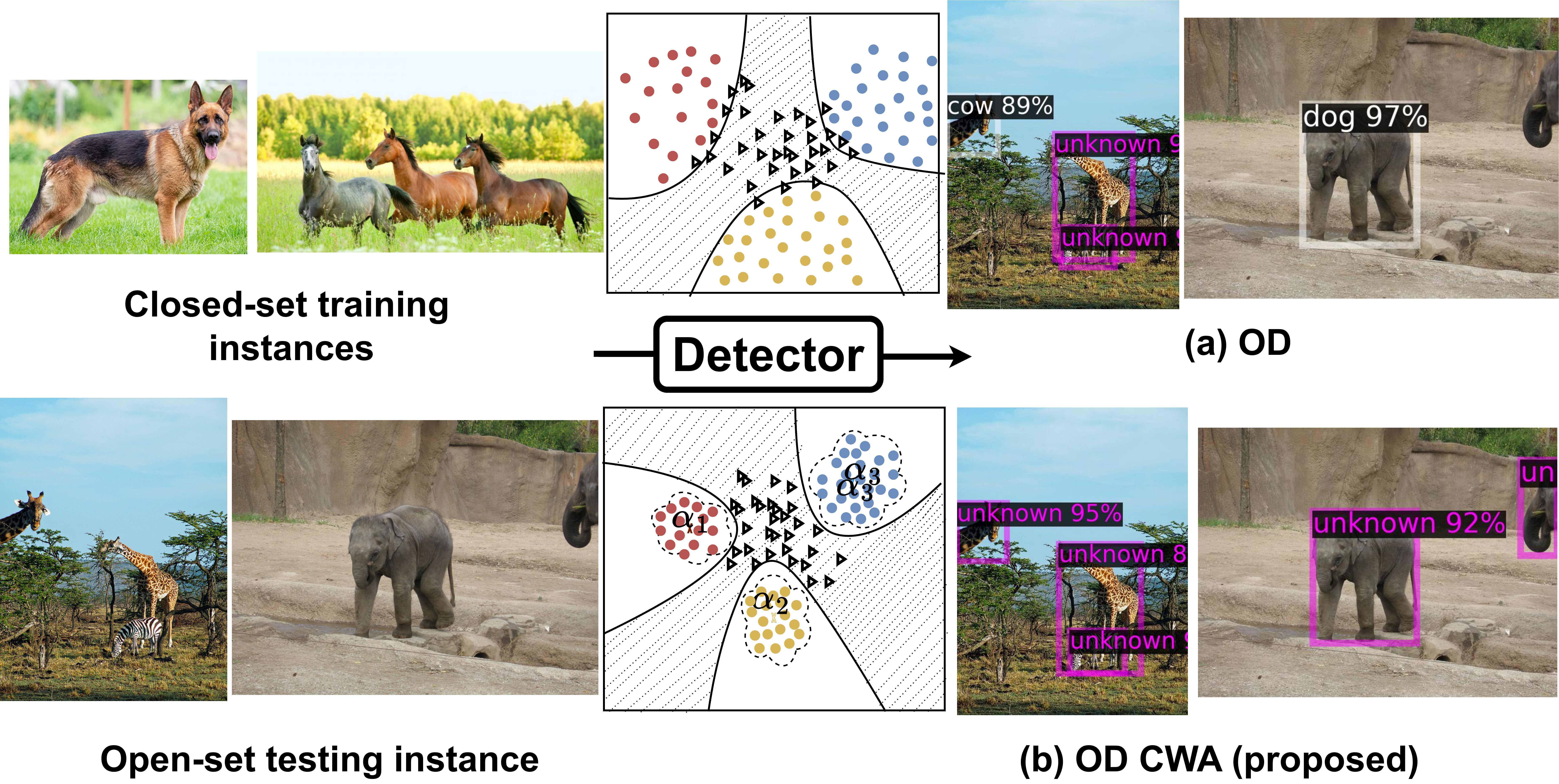}}
    \caption{{Model trained on an existing method, i.e., Open-Det (OD) (both ResNet and Transformer-based) is proficient} at identifying unknown entities to a certain extent but remains largely susceptible to misclassification of a diverse range of unfamiliar elements
(black triangles, e.g. zebra, elephant) into known classes (coloured dots, e.g.
dog, cow). (b) Our
method identifies unknown objects by enhancing the compactness among proposal features, thereby assisting the uncertainty-based optimiser in extending {low-density regions (dotted striped regions in between boundaries)} beyond the baseline (OD).}%
    \label{fig:vis_images}%
\end{figure}

In this paper, our emphasis is on an approach guided by the widely recognised principle that familiar objects tend to aggregate, creating high-density regions within the latent space. Conversely, unknown objects or novel patterns are commonly dispersed in low-density regions~\cite{self-supervvised_learning,NIPS2004_96f2b50b}. Notably, Han~\etal~\cite{han2022opendet} successfully identified unknowns without relying on complex pre-processing procedures in their work. Although this type of OSOD has improved the practicality of object detection by enabling detection of instances of unknown classes, there is still substantial room for improvement; e.g., one can refer to Fig.~\ref{fig:vis_images}, that shows detector trained using~\cite{han2022opendet} suffers from misclassifications. These open-set errors are as a result of compactness in clusters that can be further enhanced using combination of ideas from Miller~\etal~\cite{Dimity_gmm} and Courty~\etal~\cite{courty2014domain}. Additionally, motivated from Liu~\etal~\cite{spec_norm}, we leverage the distance awareness property of the model to properly quantify the distance of a testing example from the training data manifold through spectral normalisation of the weights.  

\begin{figure}%
    \centering
   { \subfloat[\centering OD]{{\includegraphics[width=3.7cm]{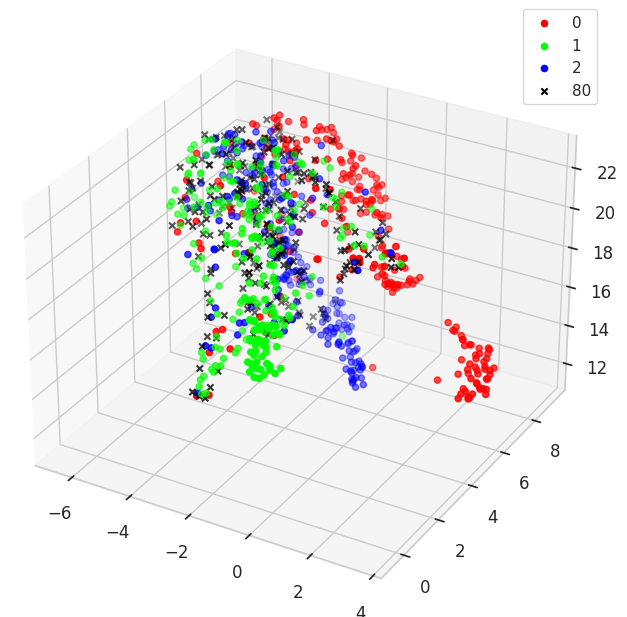}}}}
    \qquad
   {\subfloat[\centering OD-CWA]{{\includegraphics[width=3.7cm]{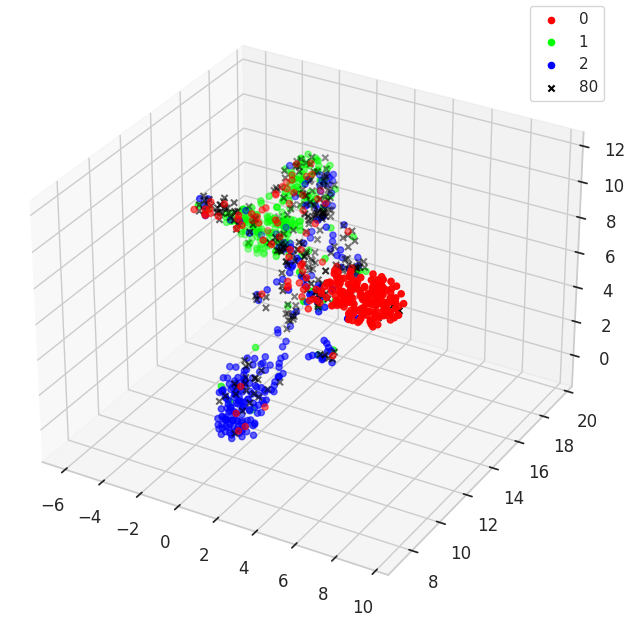} }}}%
    \caption{U-MAP~\cite{mcinnes2020umap} visualisation of latent features. Three VOC
classes (coloured dots; id - $0,1,2$) are known classes, and non-VOC classes (black cross; id - $80$) in
COCO as unknown classes. OD-CWA exhibits better separation as compared to OD in-terms of both open-set and closed-set classes.}%
    \label{fig:fi1}%
\end{figure}

In brief, the features for known classes are extracted and compactness of those features are enhanced, which in turn dictates more low-density regions for unknown classes. Then, an unknown probability as per Han~\etal~\cite{han2022opendet} is determined for each instance that serves as a threshold mechanism to distinguish low-density regions surrounding the clusters of known classes. One can visualise the latent features\footnote{Please note that we only show a small subset of classes as it provides a better visualisation of known.} for three VOC classes (coloured dots), which are known classes, and non-VOC classes (black cross) in
MS-COCO that are unknown classes; see e.g., Fig.~\ref{fig:fi1}. 

The contributions of this research can be summarised as follows, with references to relevant works:
\begin{itemize}

\item We develop a new approach, called OD-CWA, based on the Wasserstein distance (optimal transport~\cite{shen2018wasserstein,courty2014domain}) within the framework of metric learning. To the best of our knowledge, Wasserstein distance is utilised here for the first time to demonstrate OSOD. As a result of this novel setting, we demonstrate its capability in yielding promising outcomes, fostering an enhanced distinguish-ability between the known and unknown classes including flagging of unknown classes.  

\item We leverage the concept of spectral normalisation into the output linear layer that leads to the improvement of the overall effectiveness of the approaches.

\item The approach OD-CWA when compared to previous methods, exhibits significant improvements on four open-set metrics when evaluated on three different datasets, e.g., OD-CWA reduces absolute open-set errors by $17\%-22\%$ and increases the novelty detection ability by $1.5\%-16\%$. Moreover, the incorporation of these new loss functions leads to enhanced intra-class and inter-class clustering compared to the baseline OD, facilitating the expansion of low-density latent regions.

\end{itemize}

	The paper is organised as follows: Section~\ref{background} delves into open-set object detection, providing an overview of the existing work, background information, and the motivation driving the development of open-set detectors. The subsequent section~\ref{sec:PS} describes the underlying problem. Following which, Section~\ref{Wasserstein_section} provides understanding of a peculiar mathematical concept, how and where we fit it in this framework of open-set detection. Next, section~\ref{methodlogy} throws light on the setup and brief description of this novel approach. Following which, Section~\ref{sec:experimental_eval} carries out an extensive experimental evaluation of this approach on detection metrics. Then, Section~\ref{sec:conclusion} will summarise the research, highlighting certain limitations, and suggesting potential avenues for future work.
\begin{figure*}[t]
	\centering
	\setlength{\fboxsep}{3pt} 
	\fbox{\includegraphics[width=0.948\linewidth]{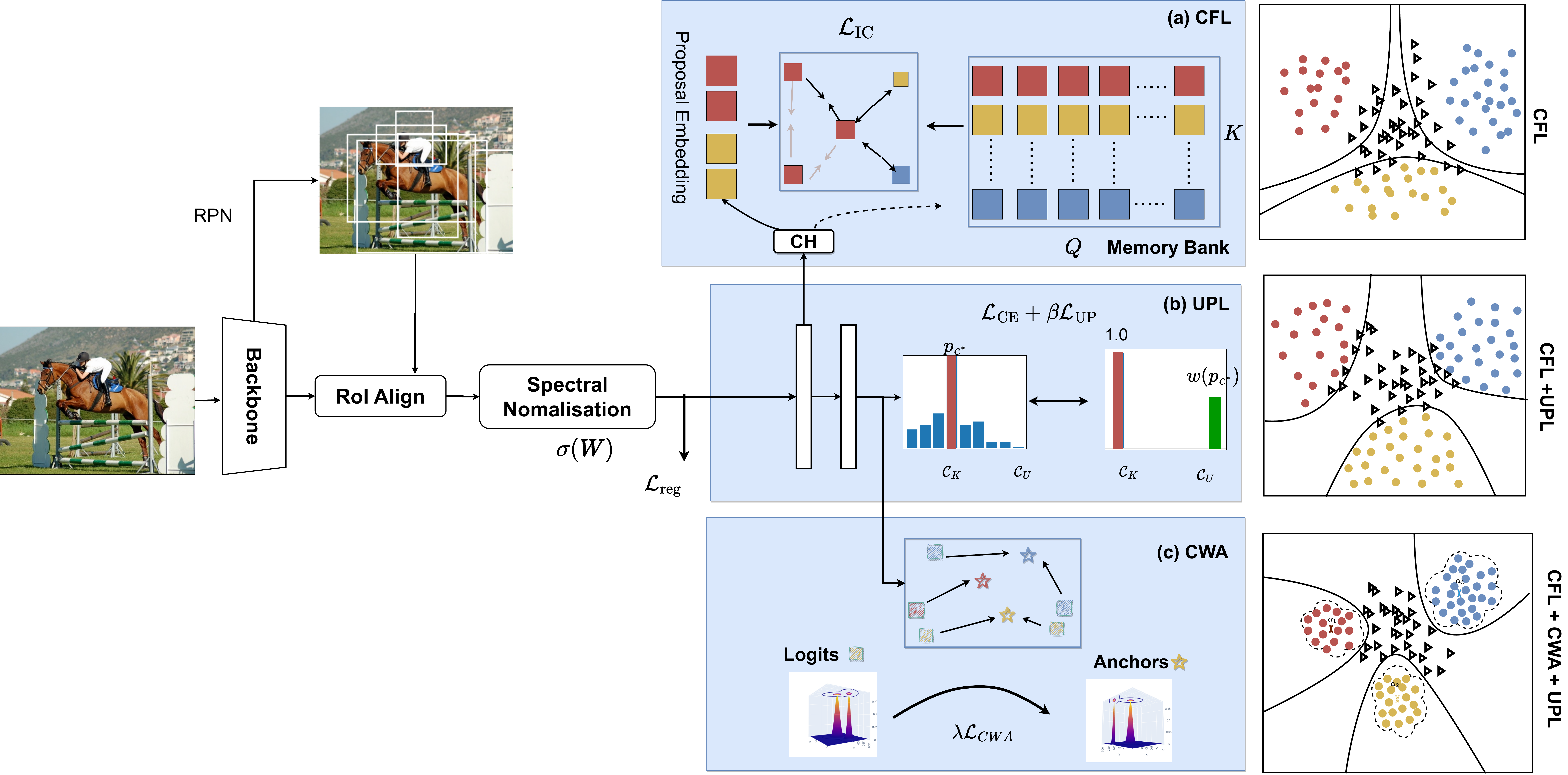}}
	\caption{OD-CWA consists of a Contrastive Feature Learner (CFL), Spectral Normalisation (SN), Class Wasserstein Anchor Learner (CWA) and Unknown Probability Learner (UPL). The CFL~\cite{han2022opendet} components utilises proposal features encoded into low-dimensional embeddings using the Contrastive Head (CH) optimised using Instance Contrastive Loss ($\mathcal{L}_{IC}$). The weights of the linear output layer are passed through a spectral normalisation step that maintain distance awareness property. Then UPL component utilises the cosine distances between embeddings and spectral normalised weights to learn the probabilities for both known classes ($C_K$) and the unknown class ($C_U$). The class Wasserstein anchor($\mathcal{L}_{CWA}$) part aids both CFL $\&$ UPL to increase the compactness in the clusters by finding the optimal transport plan from the logit space to anchor space. There is a visual illustration exhibiting the working of different components. Coloured dots and triangles represent reduced dimension of proposal features of different known and unknown classes, respectively. Coloured square represents proposal embeddings, and coloured $+$ sketched squares inside CWA box represents scaled and transformed logits.  
}
	\label{fig:cwa}
\end{figure*}
\section{Background and Related Work}\label{background}
\textbf{Open-set Object Detection: }Initial endeavours by Scheirer~\etal~\cite{scheirer2013} pioneered the investigation of open-set recognition, addressing incomplete knowledge during training where previously unseen classes may arise during testing. They devised a one-vs-rest classifier to identify and reject samples from unknown classes. Subsequently, Scheirer\etal~\cite{scheirer2014}, have expanded upon their foundational framework to further enhance the capabilities and performance of open-set recognition. Further, research by~\cite{Dimity_gmm,Dimity_dropout} has provided great insights into the utilisation of label uncertainty in open-set object detection using dropout sampling (DS) and leveraging the power of Gaussian mixture models (GMM) to capture likelihoods for rejecting open-set error rates, which are required for safety-critical applications. The complex post-processing steps involved with fitting GMMs, coupled with the reduced dispersion of low-density latent regions, lead to confident open-set errors that impede its usage in practical applications. An additional array of prevalent techniques for extracting epistemic uncertainty hinge on resource-intensive sampling-based methodologies, such as Monte Carlo (MC) Dropout~\cite{gal2016dropout} and Deep Ensembles~\cite{lakshminarayanan2017simple}. These approaches, although holding promise in the realm of OSOD~\cite{Dimity_dropout,lakshminarayanan2017simple}, carry a notable computational burden as they necessitate multiple inference passes for each image.

\textbf{Open World Object Detection and Grounding DINO}

In the context of object detection, \enquote{Open World Object Detection} (ORE) has gained prominence~\cite{OpenWorld}. This paradigm requires models to perform two crucial tasks: identifying previously unknown objects without prior supervision and continually learning these new categories without forgetting previously learned ones. To address these two tasks, ORE utilises contrastive clustering and energy-based methods to identify and integrate unknown objects. Similar, albeit with a slightly different purpose, a recent methodology called Grounding DINO~\cite{liu2023grounding} has been introduced. This method involves the integration of the Transformer-based detector DINO~\cite{zhang2022dino} with grounded pre-training techniques, which empowers the detector to identify myriad objects based on category inputs by humans, allowing it to identify pre-defined object categories. In this approach, language is introduced as a means to imbue a traditionally closed-set detector with the capability for open-set concept generalisation. This particular concept is high on generalisation, moderate on adaptability (since it's not focused on learning over time but could be adapted to), and moderate on recognition scope since it deals with a broad range of categories but doesn't necessarily focus on detecting unknown unknowns. 
\begin{figure}[t]
    \centering
{ \subfloat[\centering OD]{{\includegraphics[width=6.2cm]{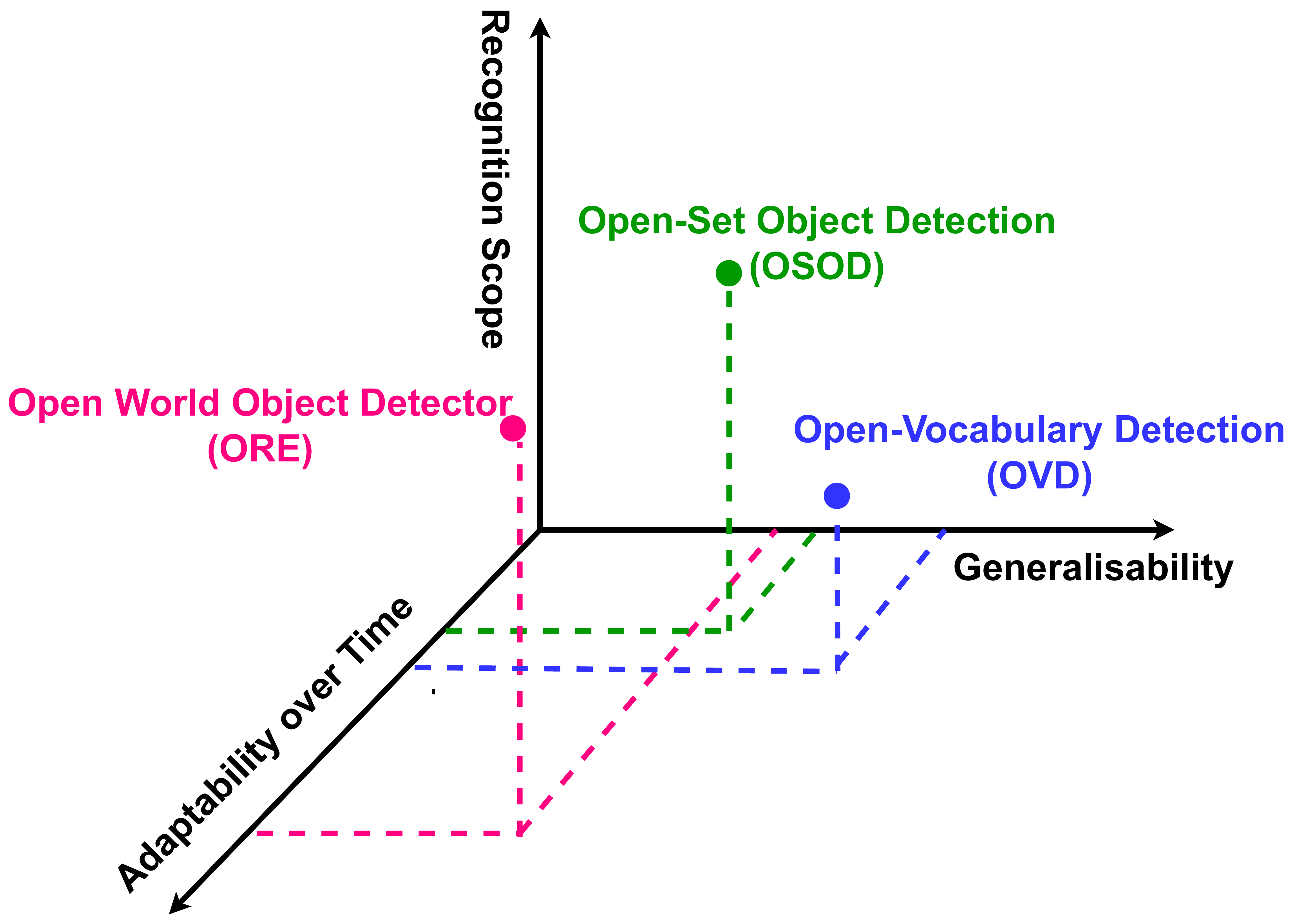}}}}
    \caption{Comparing ORE, OSOD and OVD in terms of generalizability, adaptability over time and scope of recognition.}%
    \label{fig:holistic_view_osod_ore_ovd}%
\end{figure}
On the other hand, OSOD would be positioned with moderate generalisation, low adaptability, and broad recognition scope. 
{We perceive the terms OSOD~\cite{han2022opendet,zheng2022openset}, ORE~\cite{OpenWorld}, and open-vocabulary object detection (OVD)~\cite{zareian2021openvocabulary} as inherently distinct, despite some overlap. Collectively, they can be conceptualised to exist along a spectrum encompassing \textit{recognition score, adaptability over time}, and \textit{generalizability}, as illustrated in Fig.~\ref{fig:holistic_view_osod_ore_ovd}.
Generalisation capability represents the model's ability to generalise beyond its training data, from specific known categories (low generalisation) to a wide range of categories described in natural language (high generalisation).
Adaptability over time denotes a model's ability to learn and incorporate new information over time, ranging from static models that do not learn from new data (low adaptability) to dynamic models that continuously update themselves with new categories (high adaptability). Recognition scope reflects the scope of recognition, from recognising only known classes (narrow scope) to detecting and acknowledging the presence of completely unseen classes (broad scope).} Our proposed method lies in the spectrum surrounding the point denoted by green colour.


\section{Open-Set Detection Problem Statement}  \label{sec:PS}
In an open-set detection task, one needs to detect the classes that are \textbf{seen/known} by the model (denoted as $C_\mathcal{K}$) and also flag the classes that are unseen by the model (denoted as $C_\mathcal{U}$). Let's denote the training  samples $ \mathcal{D}_{\text{closed}} := \{\mathbf{Z}_i = (\mathbf{x}_i,y_i)\}_{i=1}^{N_{\text{C}}}$, where $i$ is an instance, and is drawn from a product space $\mathcal{Z}_k = \mathcal{X} \times \mathcal{Y}_k$, where $\mathcal{X}$ and $\mathcal{Y}_k$ are the input space and label space respectively. 
The following is the underlying problem which this paper is trying to solve/deliver:
\begin{itemize}
	\item Detector to classify and detect correctly a closed set class i.e., to devise a classifier $h = g \circ f $, where $h$ is the composition of functions (likely hidden residual blocks), where \( f \) is the output of the initial operation, that will lead to features and \( g \) processes that output further to deliver logits. This is done in a way that optimises the standard objective:	
	$ \qquad \frac{1}{N_C} \sum_{i=1}^{N_C}  \Big[ y_i = \arg \max_{c\in \mathcal{Y}_C} f(\mathbf{x_i})_{y_i}) \Big],$ where $N_C$ is the number of closed set samples and $C$ refers to the total number of known classes.

\item Detector to identify the open-set detections  and flag them into a broader class of unknown using some form of score function.The score function can take various formulations, and we employ a specific unknown optimisation approach from the literature, detailed in subsequent sections.
\end{itemize}
\section{OSOD as an optimal-transportation problem} \label{Wasserstein_section}
Due to the prominence of Wasserstein distance in representational learning, domain adaptation~\cite{courty2014domain,shen2018wasserstein} and generative adversarial networks~\cite{pmlr-v70-arjovsky17a} we leverage this ideology and adapt it to the OSOD problem. Before diving into the details of how to pose Wasserstein distance inside OSOD, we first provide a detailed understanding of the Wasserstein metric. 

\textbf{Wasserstein Distance} is a distance
measure between probability distributions on a given metric space $(M,\rho)$, where $\rho(x,y)$ is a distance function for two instances $x$ and $y$. The $p$-th Wasserstein distance between two Borel probability measures\footnote{The definition of measurable space can be found~\cite[Page 160]{Billingsley1995probability}} $\mathbf{P}$ and $\mathbf{Q}$ is defined as,
\begin{definition}[Shen~\etal~\cite{shen2018wasserstein}]
For $p\geq1$, the \textit{p-}Wasserstein distance between two distribution $\mathbf{P} \text{ and } \mathbf{Q}$ is given by, 
    \begin{align} \label{eq:wass_dist}
	\mathcal{W}(\mathbf{P},\mathbf{Q}) & \triangleq \Bigg[ \inf_{\theta } \int_{X \times X} d(x,y)^p , d\theta(x,y) \Bigg]^{\frac{1}{p}}
\end{align}
\end{definition}
The term ${d(x,y)}^p= ||x-y||^{p}$ represents the cost and $\mathbf{P}, \mathbf{Q} \in \{\mathcal{P} :  \int {d(x, y)}^p \, d \mathbf{P}(x) < \infty,  \forall y \in M$ are two probability measures with finite $p$-th moment. This metric fundamentally arises in the optimal transport problem, where $\theta(x, y)$ is a policy for transporting one unit quantity from location $x$ to location $y$ while satisfying the constraint $x \sim \mathbf{P}$ and $y \sim \mathbf{Q}$. When the cost of transporting a unit of material from $x \in \mathbf{P}$ to $y \in \mathbf{Q}$ is given by $d(x, y)^{p}$, then $\mathcal{W}(\mathbf{P}, \mathbf{Q})$ is the minimum-expected transport cost. Please note that solving $\mathcal{W}(\mathbf{P},\mathbf{Q})$ is a challenge and therefore pioneering work according to the Kantorovich-Rubinstein theorem deals with the dual representation of the first Wasserstein distance that can be expressed as an integral probability metric \cite{OT}:
\begin{equation}
\mathcal{W}_1(\mathbf{P}, \mathbf{Q}) = \sup_{\|f\|_{\text{Lip}} \leq 1} \left( \mathbb{E}_{x \sim \mathbf{P}}[f(x)] - \mathbb{E}_{x \sim \mathbf{Q}}[f(x)] \right),  \label{eq:mainwass}
\end{equation}
where the Lipschitz semi-norm is defined as $\|f\|_{\text{Lip}} = \sup \left|f(x) - f(y)\right|/ d(x, y)$. In order to make a tractable computation of $\mathcal{W}_1(\mathbf{P}, \mathbf{Q})$, an algorithm\footnote{Details can be found in Ramdas~\etal\cite{ramdas2015wasserstein}.} called as Sinkhorn-Knopp~\cite{Knight} algorithm is utilised. This is an iterative algorithm used to compute an approximate optimal transport plan $T_{k, l}$ for two probability distributions ($\mathbf{P} \text{ and } \mathbf{Q}$) while applying a regularisation as per the equation below,
\begin{equation} \label{eq:sinkhorn}
 D_{\text{SH}} (T; \mathbf{P}, \mathbf{Q}) = \sum_{k,l} T_{k, l}   d_{k,l}(x,y) -  \rho \cdot \sum_{k,l} T_{k, l}  \log(T_{k,l}))
\end{equation}
where $\rho$ is smoothing parameter and $k,l$ are elements (can be matrix variate) sampled from $\mathbf{P}$ and $\mathbf{Q}$. 
\begin{remark}
  In the context of the paper, the distribution of $\mathbf{P}$ would refer to the distribution of anchors $\mathbf{A} \in \mathbb{R}^{B\times K \times K}$ and $\mathbf{Q}\in \mathbb{R}^{B\times K \times K}$ is the distribution of logits $\mathbf{L}$, where $B = 512 \times b$, where $b$ is the batch size, and $K$ is the total number of known classes. The class anchor clustering by~\cite{dimityCAC} takes into account $||\mathbf{L}-\mathbf{A}||_{2}$. However, in a higher-dimensional space, namely $\mathbb{R}^{B \times K \times K}$, the estimation of parameters for deep neural networks that optimally transfer the samples of distributions of two probability spaces ($\mathbf{L}$ and $\mathbf{A}$) can be more effectively addressed through the lens of optimal transport, as encapsulated by the Wasserstein distance~\eqref{eq:mainwass}. This concept serves as the cornerstone of this paper.  
\end{remark}
\begin{figure*}%
    \centering
     {\includegraphics[width=17.4cm]{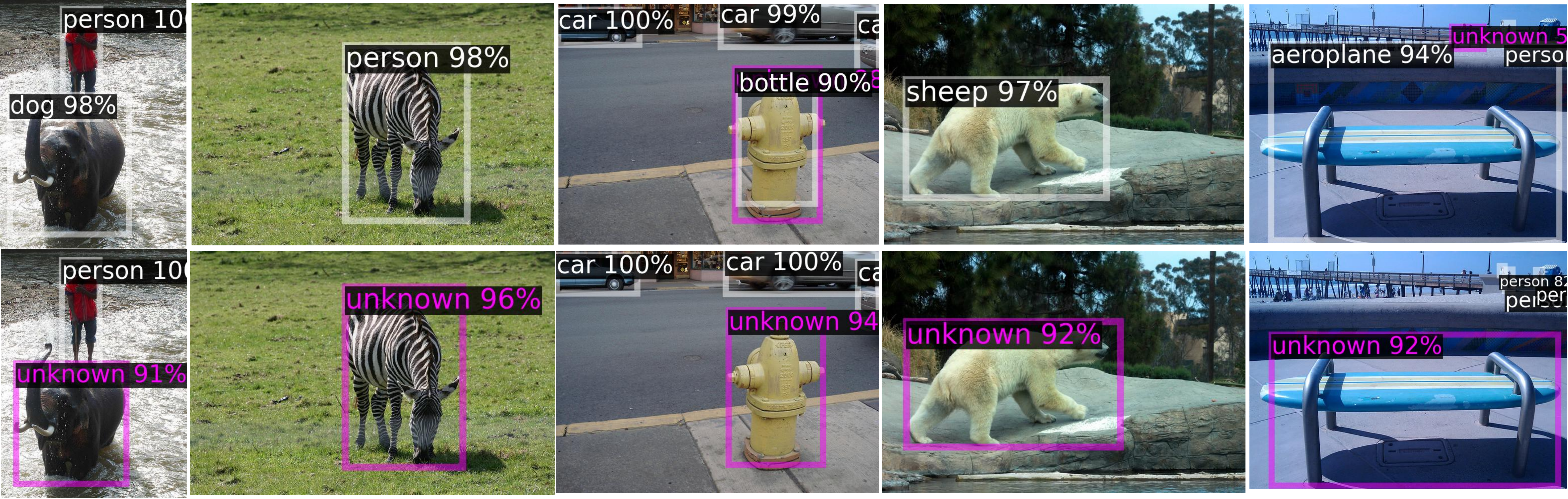}}
    \caption{Qualitative comparisons between proposed OD (top) and OD-CWA (bottom). Both models are trained on VOC and the detection results are visualised using images from COCO. The purple colour represents unknown and white represents known. White annotations represent classes seen by the model and purple annotation correspond to unknown classes.}%
    \label{fig:inference_images}%
\end{figure*}
\section{Methodology} \label{methodlogy}

We employ the Faster R-CNN~\cite{fasterrcnn} framework consisting of a backbone, Region Proposal Network (RPN), and region-convolutional neural network (R-CNN) that follows the architecture by Han~\etal~\cite{han2022opendet}. We augment the cosine similarity-based classifier by introducing a spectral normalisation of the weights to the final linear layer. Some of the details are delineated below.
\subsection{Spectral Normalisation of Weights to Linear Layer} \label{subsec:SN}
This specific configuration draws inspiration from Liu et al.~\cite{spec_norm} which explains that deep learning models, e.g., ResNets, Transformers, often adopt residual architecture blocks, represented as $h(\mathbf{x}) = h_{L-1} \circ \ldots \circ h_2 \circ h_1(\mathbf{x})$, where $h_l(\mathbf{x}) = g \circ f$ refers to hidden units of layer $l$. They also mention that hidden mapping can be made to have meaningful relationship to the distance in input space. This is called distance preserving property that quantifies the distance from training data manifold and can be carried out by constraining Lipschitz constants of layers in the residual mappings $g_l$. We utilise this concept to ensure that the weight matrices linked with linear final residual block $g_1(\mathbf{x})$ has a spectral norm, i.e., the largest singular value as less than 1 and is due to the inequality $\|g_l\|_{\text{Lip}} \leq \|W_l \mathbf{x} + b_l\|_{\text{Lip}} \leq \|W_l\|_2 \leq 1$, where $l=1$. Therefore, we employ spectral normalisation (SN) on the weight matrices, $\{W_l\}_{l=1}$ i.e., last layer. Subsequently, the spectral normalised weights are utilised in conjunction with the features within a scaled cosine similarity scoring function, producing output logits as follows: $\mathbf{L}_{i,j} = \frac{\alpha \cdot \mathbf{F}(\mathbf{x})_i^T \cdot \mathbf{w}_j}{\|\mathbf{F}(\mathbf{x})_i\| \cdot \|\mathbf{w}_j\|} \in \mathbb{R}^{m \times C}$, where $\mathbf{L}_{i,j}$ signifies the similarity score between the $i$-th proposal features $\mathbf{F}(\mathbf{x})_i$ and the weight vector $\mathbf{w}_j$ of class $j$, $m$ denotes the multi-level features, and $C$ is the total number of known classes. The scaling factor $\alpha$ is set to $20$, and the box regressor is configured to be class-agnostic as per Han et al.~\cite{han2022opendet}.
\subsection{Supervised Contrastive Loss}
The primary purpose of this component of the loss is to create compactness between detections of individual classes (intra-class). This compactness and separation ultimately leads to expansion of low-density latent regions by narrowing the cluster of known classes. Motivated from Han~\etal~\cite{han2022opendet} i.e., CFL contains a contrastive head (CH), a memory bank and an instance level contrastive loss function $\mathcal{L}_{IC}$ and each proposal feature  $ \mathbf{F}(\mathbf{x})_i $ is encoded into an embedding with CH, that is optimised from the memory bank with the help of $\mathcal{L}_{IC}$. The formulation of $\mathcal{L}_{IC}$ is described by, $ \mathcal{L}_{IC}  = \frac{1}{N} \sum_{i=1}^N \mathcal{L}_{IC} (\mathbf{z}_i)$, where,
\textcolor{black}{
\begin{align*}
\mathcal{L}_{IC} (\mathbf{z}_i)  =    \frac{-1}{|M(c_i)|} & \sum_{\mathbf{z}_j \in M(c_i)} \log  \frac{\exp ( \frac{\mathbf{z_i} \cdot  \mathbf{z_j}}{\tau})}{\sum_{\mathbf{z}_k \in A(c_i)} \exp (\frac{\mathbf{z_i} \cdot  \mathbf{z_k}}{\tau}) }   , 
\end{align*}
}
and $\mathbf{z}_i$ corresponds to the embeddings of $i^{\text{th}}$ proposal and $\tau$ is the temperature parameter. For detail about this loss function, one can follow Khosla~\etal~\cite{supcont}. 
\subsection{Class Wasserstein Anchor Loss}
This approach is motivated from Miller~\etal\cite{dimityCAC} and Wen~\etal~\cite{wen2016discriminative} that utilises the centre loss function as per,
\begin{align}
    \mathcal{L}_{CWA} = \sum_{i=1}^{m} \mathcal{W}(\mathbf{L}_i,\mathbf{A}_{y_i}),
\end{align}
where $\mathbf{A}_{y_i} \in $ is the class anchor centres of deep features corresponding to $y^{th}$ class and  $\mathbf{L}_i$ denotes the spectral normalised scaled logits as mentioned in Section~\ref{subsec:SN}. The function $\mathcal{W}$  is computed as per Eq.~\eqref{eq:wass_dist} and approximated as per Eq.~\eqref{eq:sinkhorn} using iterative methods laid out in the implementation section.
\subsection{Unknown Probability Loss}
Unknown Probability Loss is an integral part of the primary loss function, which works in conjunction with CFL and CWA to enhance the scattering of the low-density regions around the cluster of known classes. The SN-transformed logits, $\mathbf{L}_i$ of a proposal are passed through a softmax function that leads to a softmax cross entropy (CE) loss $\mathcal{L}_{CE} = \sum_{c \in C} y_c \log(p_c)$, where $y_c = 1$ when $c = c^t$ is a ground truth class, $p_c = \frac{ \exp (\mathbf{L}_c)}{\sum_{j \in C} \exp (\mathbf{L}_j)}$ and $C=C_{\mathcal{K} \cup \mathcal{U} \cup \text{bg} }$ that denotes all known, unknown inclusive of background. This CE loss along with a weighted formulation of entropy that acts as an uncertainty signal is utilised for constructing the unknown probability loss function $\mathcal{L}_{UP} \triangleq - w(p_{c^t}) \log (\bar{p}_{\mathcal{U}}) )$, where $\bar{p}_\mathcal{U} \triangleq \frac{ \exp (\mathbf{L}_c)}{\sum_{j \in C \setminus \{c^t\}} \exp (\mathbf{L}_j)} $ is the softmax without the logit of the ground truth class ${c^t}$. The term, $w(p_{c^t}) = (1-p_{c^t})^\alpha p_{c^t}$. Further details about optimising this function can be found in Han~\etal~\cite{han2022opendet}. 

\subsection{Optimisation Objective Function}
The combined loss function employed in this paper can be trained in an end-to-end manner as follows:
\begin{align}
    \mathcal{L}_{\text{OD-CWA}}  & = \bar{\mathcal{L}}    + \sigma \big[\lambda  \mathcal{L}_{CWA}+ \beta \mathcal{L}_{UP} + \mathcal{L}_{CE} \big] + \delta_k \mathcal{L}_{IC},  \label{full_equation} \\
        \mathcal{L}_{\text{OD-SN}}  & = \bar{\mathcal{L}}    + \sigma \big[\beta \mathcal{L}_{UP} + \mathcal{L}_{CE} \big] + \delta_k \mathcal{L}_{IC}, \label{eq:od-sn}
\end{align}
 {where}, $\bar{\mathcal{L}} = \mathcal{L}_{rpn} + \mathcal{L}_{reg}$, $\sigma$ denotes SN component, $\mathcal{L}_{CWA} =  \mathcal{W}_1(\mathbf{P}, \mathbf{Q})$, \text{ as per \eqref{eq:mainwass}} and evaluated in an iterative manner based on eq~\eqref{eq:sinkhorn} and $\mathcal{L}_{\text{rpn}}$ denotes the total loss of RPN, $\mathcal{L}_{\text{reg}}$ is the smooth $\text{L}_1$ loss for box regression, $\beta$, $\delta_k$ and $\lambda$ are weighting coefficients of UPL, CFL and CWA parts of the loss function. The parameter $\delta_k$ is proportional to the current iteration $k$ and is set to gradually decrease weight of $\mathcal{L}_{IC}$~\cite{han2022opendet}. The parameter
$\gamma$ dictates the impact of the Wasserstein distance-based loss, $\mathcal{L}_{CWA}$ which is sensitive to its impact on the overall objective. A comprehensive depiction of the OD-CWA framework is presented in detail in Fig.~\ref{fig:cwa}.
\begin{table*}[htbp]
	\small
 \centering
	\setlength{\tabcolsep}{4pt}
	\renewcommand{\arraystretch}{1.12}
	\begin{tabular} {@{}c|@{}c|c@{}@{}c@{}c@{}@{}c|c@{}@{}c@{}c@{}@{}c|c@{}@{}c@{}c@{}@{}c}
		\hline
		Method & VOC & \multicolumn{4}{c|}{VOC-COCO-20} & \multicolumn{4}{c|}{VOC-COCO-40} & \multicolumn{4}{c}{VOC-COCO-60} \\
		\cline{2-5} \cline{6-9} \cline{10-14}
		& $mAP_{K\uparrow}$ & $WI_{\downarrow}$ & $AOSE_{\downarrow}$ & $mAP_{K\uparrow}$ & $AP_{U\uparrow}$ & $WI_{\downarrow}$ & $AOSE_{\downarrow}$ & $mAP_{K\uparrow}$  & $AP_{U\uparrow}$ &  $WI_{\downarrow}$ & $AOSE_{\downarrow}$ & $mAP_{K\uparrow}$ & $AP_{U\uparrow}$ \\
		\hline
		FR-CNN~\cite{Large-ScaleLong-Tailed}      & \textbf{80.10} & 18.39 & 15118 & \textcolor{orange}{58.45} & - & 22.74 & 23391 & 55.26 & - & 18.49 & 25472 & 55.83 & - \\
  		CAC~\cite{Dimity_gmm}      & 79.70 & 19.99 & 16033 & 57.76  & - & 24.72 & 25274 & 55.04 & - &  20.21 & 27397 & 55.96 & - \\
		PROSER~\cite{zhou2021learning}  &   79.68 & 19.16 &13035 &57.66 & 10.92 & 24.15 & 19831 & 54.66 & 7.62 &19.64  & 21322& 55.20 & 3.25 \\
		ORE~\cite{OpenWorld}       &  79.80 & 18.18 & 12811 & 58.25 & 2.60 & 22.40 & 19752 & 55.30 & 1.70 & 18.35 & 21415 & 55.47 & 0.53 \\
		DS~\cite{Dimity_dropout}           & \textcolor{orange}{80.04} & 16.98 & 12868 & 58.35 & 5.13 & 20.86 & 19775 & 55.31 & 3.39 & 17.22 & 21921 & 55.77 & 1.25\\
		 {OD}~\cite{han2022opendet}   & {80.02} & 14.95 & 11286 & \textbf{58.75} & 14.93 & 18.23 & 16800 & \textbf{55.83} & \textcolor{orange}{10.58} & 14.24 & 18250 & \textbf{56.37} & \textcolor{orange}{4.36} \\
   \hline
		{OD-SN} & {79.66} & \textcolor{orange}{12.96} & \textcolor{orange}{9432} & {57.86} & \textcolor{orange}{14.78} & \textcolor{orange}{16.28} & \textcolor{orange}{14118} & \textcolor{orange}{55.36} & {10.54} & \textcolor{orange}{12.76} & \textcolor{orange}{15251} & \textcolor{orange}{56.07} & {4.17} \\
	{OD-CWA} & {79.20} & \textbf{11.70} & \textbf{8748} & {57.58} & \textbf{ 15.36} & \textbf{ 14.58} & \textbf{13037} & {55.26} & \textbf{10.98} & \textbf{11.55} & \textbf{14984} & {55.73} & \textbf{4.45} \\
		\hline
	\end{tabular}
\caption{Comparison of OD-SN and OD-CWA (trained on ResNet-50 backbone) with other methods on VOC and VOC-COCO-$\{\textbf{T}_1\}$. The close-set performance ($mAP_{K}$) on VOC, and both close-set ($mAP_{K}$) and open-set ($WI, AOSE, AP_\mathcal{U}$) performance of different methods on VOC-COCO-20, 40, 60 are reported. Numbers in bold black colour indicates best performing on that metric, and bold orange indicates second best. Significant improvements in $WI, AOSE$ and $AP_\mathcal{U}$ are achieved at the expense of a slight decrease in $mAP_{\mathcal{K}}$.}
	\label{tab:object-detection-performance-r50_coco_voc}
\end{table*}

\begin{table*}[htbp]
	\small
 \centering
	\setlength{\tabcolsep}{4pt}
	\renewcommand{\arraystretch}{1.12}
	\begin{tabular} {@{}c|c@{}@{}c@{}c@{}@{}c|c@{}@{}c@{}c@{}@{}c|c@{}@{}c@{}c@{}@{}c}
		\hline
		Method &  \multicolumn{4}{c|}{VOC-COCO-0.5n} & \multicolumn{4}{c|}{VOC-COCO-n} & \multicolumn{4}{c}{VOC-COCO-4n} \\
		\cline{2-5} \cline{6-9} \cline{10-13}
		&  $WI_{\downarrow}$ & $AOSE_{\downarrow}$ & $mAP_{K\uparrow}$ & $AP_{U\uparrow}$ & $WI_{\downarrow}$ & $AOSE_{\downarrow}$ & $mAP_{K\uparrow}$  & $AP_{U\uparrow}$ &  $WI_{\downarrow}$ & $AOSE_{\downarrow}$ & $mAP_{K\uparrow}$ & $AP_{U\uparrow}$ \\
		\hline
		FR-CNN~\cite{Large-ScaleLong-Tailed}    &  9.25 & 6015 & 77.97 & - & 16.14 & 12409 & 74.52 & - & 32.89 & 48618 & 63.92 & - \\
  		CAC~\cite{Dimity_gmm}      & 9.92  & 6332 & 77.90 & - &  16.93 & 13114 &  74.40 &  - & 35.42 & 52425 & 63.99 & - \\
		PROSER~\cite{zhou2021learning}  &   9.32 & 5105 & 77.35 & 7.48 & 16.65 & 10601 & 73.55 & 8.88 & 34.60 & 41569 & 63.09 & 11.15 \\
		ORE~\cite{OpenWorld}    &  8.39 & 4945 & 77.84 & 1.75 & 15.36 & 10568 & 74.34 & 1.81 & 32.40 & 40865 & 64.59 & 2.14 \\
		DS~\cite{Dimity_dropout}    &   8.30 & 4862 & 77.78 & 2.89 & 15.43 & 10136 & 73.67 & 4.11 & 31.79 & 39388 & 63.12 & 5.64 \\
		{OD}~\cite{han2022opendet}   & 6.44 & 3944 & \textbf{78.61} & \textbf{9.05} & 11.70 & 8282 & \textbf{75.56} & \textbf{12.30} & 26.69 & 32419 & \textbf{65.55} & \textcolor{orange}{16.76} \\
  \hline
{OD-SN}   & \textcolor{orange}{6.01} & \textcolor{orange}{3084} & {78.16} & {8.24} & \textcolor{orange}{11.30} & \textcolor{orange}{6439} & {75.11} & {11.92} & \textcolor{orange}{26.50} & \textcolor{orange}{26261} & {64.49} & { 16.48} \\
{OD-CWA}   & \textbf{5.21} & \textbf{2780} & \textcolor{orange}{78.30} & \textcolor{orange}{8.30 } & \textbf{ 9.95} & \textbf{6001} & \textcolor{orange}{75.10} & \textcolor{orange}{12.26 } & \textbf{23.31 } & \textbf{24072} & \textcolor{orange}{64.75} & \textbf{ 17.11 } \\
		\hline
	\end{tabular}
	\caption{Comparison of OD-SN and OD-CWA (trained on ResNet-50 backbone) with other methods on VOC and VOC-COCO-$\textbf{T}_2$. Numbers in bold black colour indicates improvement of that metric, and bold orange indicates deterioration of that metric. Numbers in bold black colour indicates best performing on that metric, and bold orange indicates second best.}
	\label{tab:object-detection-performance-r50_coco_voc_0.5m}
\end{table*}
\begin{table}[htbp]
    \small
    \setlength{\tabcolsep}{1.0pt}
    \renewcommand{\arraystretch}{1.05}
    \begin{tabular} {@{}c|c|@{}@{}c|c|@{}@{}cc@{}@{}cc}
        \hline
        Method & $\delta_k$ & $\lambda$ & VOC  & \multicolumn{4}{c}{VOC-COCO-20}  \\
        \cline{4-8} 
        &  & & $mAP_{K\uparrow}$ & $WI_{\downarrow}$ & $AOSE_{\downarrow}$ & $mAP_{K\uparrow}$ & $AP_{U\uparrow}$ \\
        \hline
         {OD}~\cite{han2022opendet} & 0.1  & -  & {83.29}                   &   {12.51}           &   {9875}            &   {63.17}             &   {15.77}         \\
       \hline
        OD-SN & 0.1 & - &  {82.49} &  {14.39} & \textbf{7306} & {61.59} & \textcolor{orange}{16.45}         \\
        OD-CWA &  0.25 &  1.3 &  \textbf{83.64}         & {12.44}      & {7880}     & {63.18}      & {14.06}        \\
            OD-CWA  & 0.18  & 1.3 & {83.06}         & \textbf{10.32}      & \textcolor{orange}{8888}     & {62.93}      & {15.34}       \\
        OD-CWA & 0.21 & 1.6 &   {79.20} & {12.16} & {10007} & {57.09} & {15.53}   \\
            OD-CWA  & 0.21  & 1.83 & {83.15}         & {10.44}      & {8951}     & \textcolor{orange}{63.20}      & {15.41}       \\
       OD-CWA  &   {0.21}  & {1.7} & \textcolor{orange}{83.34}         & \textcolor{orange}{10.35}      & \text{8946}     & \textbf{63.59}      & \textbf{18.22}     \\
        \hline
    \end{tabular}
    \caption{Comparison of the Swin-Transformer~\cite{swintr} backbone-based proposed method with Object Detection (OD) on VOC and VOC-COCO-$\textbf{T}_1$, examining the impacts of varying $\delta_k$ and $\lambda$ on different metrics. Parameter $\lambda$ is in the order of $10^{-3}$. OD-CWA (last row) outperforms OD in every aspect significantly.}
    \label{tab:object-detection-performance-swinTr50}
\end{table}
\section{{Experimental Evaluation}} \label{sec:experimental_eval}
\subsection{Setup}
\textbf{Datasets} 
This method utilises \texttt{PASCAL-VOC}~\cite{pascal-chsc-2011} and \texttt{MS COCO}~\cite{mscoco} dataset. First, we utilise \texttt{trainval} dataset of VOC for closed-set training, and then use a set of $20$ VOC CLASSES and $60$ non-VOC classes in COCO to evaluate the proposed method under myriad open-set conditions. Particularly, two different settings are considered, i.e., {\texttt{VOC-COCO}}-$\{\textbf{T}_1,\textbf{T}_2\}$ as per Han~\etal~\cite{han2022opendet}. We resort to similar dataset to maintain consistency. The open-set $\{\textbf{T}_1\}$, contains $5k, 10k, 15k$ VOC testing images containing $\{20,40,60\}$ non-VOC classes and  $\{\textbf{T}_2\}$  was created with four combined datasets by increasing WI~\cite{Dhamija2020TheOE}, each comprising $n=5000$ VOC testing images and disjoint sets of $\{0.5n, n, 4n\}$ COCO images not overlapping with VOC classes.
Now, we describe different metrics that are utilised for evaluation of our approach.
\textbf{Wilderness Impact (WI)~\cite{Dhamija2020TheOE}} is used to measure the degree of unknown objects misclassified to known classes and has the following formulation, $ \text{WI} = \left(\frac{P_\mathcal{K}}{P_{\mathcal{K} \cup \mathcal{U}}} - 1\right) \times 100$, where $P_{\mathcal{K}}$ and $P_{\mathcal{K} \cup \mathcal{U}}$ denote the precision of close-set and open-set classes and the metric WI is scaled by $100$ for better presentation. When it comes to evaluation, the paper takes a very similar evaluation approach as per Han~\etal~\cite{han2022opendet}. \textbf{Absolute Open-Set Error (AOSE)~\cite{Dimity_gmm}} is used, a metric used to count the number of misclassified unknown objects. Furthermore, we report the mean Average Precision (mAP) of known classes (mAP$_K$). Lastly, we measure the novelty discovery ability by $AP_{\mathcal{U}}$ (AP of the unknown class). Note WI, AOSE, and $AP_{\mathcal{U}}$ are open-set metrics, and mAP$_K$ is a close-set metric. 
\textbf{Implementation Details}
In our methodology, we employ the architecture ResNet-50~\cite{ResNet50} with a Feature Pyramid Network (FPN)~\cite{LinDGHHB17} for all our methods. Some minor changes were made to the learning rate scheduler of Detectron2~\cite{wu2019detectron2} utilised inside the framework of OD~\cite{han2022opendet}. A Stochastic Gradient Descent optimiser is used for training with an initial learning rate, momentum and weight decay to be $0.02$, $0.9$, and $10^{-4}$ respectively. Model training is carried out concurrently on $8$ GPUs (\textbf{Tesla V100-SXM2-32GB}) with $16$ images per batch. In regard to CFL, we follow the parameter settings used by OD.\footnote{All experiments are conducted using the exact training parameters outlined in Han et al.~\cite{han2022opendet}, except for our novel additions to the loss function, which ensures consistent comparison.} For UPL, we follow the uncertainty guided hard example mining procedure and select top-$3$ sample examples for both foreground and background proposals. The hyperparameters $\alpha,\beta$ are set to, $1,0.5$ respectively, $\delta_k$ is gradually decreased with every iteration $k$. For carrying out the optimal transportation, we utilise a package called as \texttt{Goemloss}~\cite{feydy2019interpolating} with tensorised PyTorch backend. The parameter \(p\) in the Sinkhorn distance loss corresponds to the power in the cost function. In our case, we choose $p=1$, indicating the utilisation of the \(L_1\) norm as the cost $d(x,y)$. The parameter $\text{blur } \nu =0.1$ introduces a smoothing factor into the optimisation process to averts numerical instability, enhancing the overall stability of the Sinkhorn distance computation. 
\subsection{Main Results}
The novelty of our paper lies in development of two main modelling approaches: Open-Det Class Wasserstein Anchor (OD-CWA) and Open-Det Spectral Normalisation (OD-SN) as per Eqs.~\ref{full_equation} and~\ref{eq:od-sn}, respectively. We compare these two approaches with $7$ other methods, which also includes the current state-of-the-art method by~\cite{han2022opendet}, on VOC-COCO-$\{\textbf{T}_1,\textbf{T}_2\}$ dataset, which are exhibited in Tables~\ref{tab:object-detection-performance-r50_coco_voc} and~\ref{tab:object-detection-performance-r50_coco_voc_0.5m}. When we train a model as per Eq.~\ref{full_equation} and evaluate for VOC and VOC-COCO-$20$ classes in $\textbf{T}_1$, we can see that there is a decrease of $ 12\%$ $\&$ $16.4\%$ for metrics WI and AOSE, respectively, while there is a $1\%$ reduction in $\text{mAP}_{\mathcal{K}}$ and $AP_\mathcal{U}$ for the ResNet-50 backbone. Similar effects can be observed in VOC-COCO-$\{40,60\}$ in $\textbf{T}_1$. On almost all metrics for $\textbf{T}_2$ dataset, similar effects of substantial improvement in WI and AOSE with slight deterioration in $\text{mAP}_{\mathcal{K}}$ and $AP_\mathcal{U}$ is observed for the ResNet-50 backbone. Further, we have added Class Anchor Clustering (CAC) to the Table~\ref{tab:object-detection-performance-r50_coco_voc} and~\ref{tab:object-detection-performance-r50_coco_voc_0.5m}, and it suffers from poor performance just by itself. The second approach, OD-CWA,  utilises SN along with the addition of $\mathcal{L}_{CWA}$, $\mathcal{L}_{IC}$ and $\mathcal{L}_{UP}$ as per Eq.~\ref{full_equation}. As a result of this combination on the ResNet-50 backbone, the improvement in open-set performance is significant, i.e., improvement of $18\%, 22 \%$ and $2.8 \%$ on open-set metrics WI, AOSE and AP$_\mathcal{U}$, respectively. The improvement can be best captured in Fig.~\ref{fig:inference_images} which depicts qualitative results comparing approaches OD-CWA and OD. Further, when we use OD-CWA approach to train a detector based on a Swin-T backbone, we can see that our detector surpasses OD by a substantial amount on all the metrics, e.g., Table~\ref{tab:object-detection-performance-swinTr50} suggests there is an improvement of $\approx 21\%, 9.4\%$, $0.6\%$ and $15.5\%$ on all $4$ metrics.
\begin{table}[htbp]
\small
	\centering
     \setlength{\tabcolsep}{1.7pt}
    \renewcommand{\arraystretch}{0.93}
	\begin{tabular}{c|c|c|c|c|c} 
		\toprule
		$\delta_k$ & $\lambda$ & $WI_{\downarrow}$ & $AOSE_{\downarrow}$ & $mAP_{K\uparrow}$ & $AP_{U\uparrow}$\\
		\midrule
0.35 & -  & 12.96 &  9432 & \textbf{57.86} & 14.78   \\
     \hline
     0.1 & 1.3  & 19.07 &  7899 & 54.94 & 14.38 \\
 		  0.35 &  2.6  & 10.46 & 9803 & 56.19 & \textcolor{orange}{15.03}    \\
		 0.35 &  4.6  & \textbf{10.40} & 9566 & 56.64 & 14.54       \\
    0.29 &  4.6  & 11.52 & \textcolor{orange}{9312} & 57.02 & 14.18   \\
     	  0.26 & 4.6   & \textcolor{orange}{11.20} & 9473 &  \textcolor{orange}{57.02} & 14.19       \\
             	$  {  0.21}$ & $ {1.7}$   & $ {11.70}$ & \textbf{8748} &  \textbf{57.86} & \textbf{15.36}        \\
		\bottomrule
	\end{tabular}
\caption{Effects of $\delta_k$ and $\lambda$ in the combined loss function for model trained using ResNet-50 backbone and evaluated on VOC-COCO-20 dataset. Parameter $\lambda$ is in the order of $10^{-3}$.  The first row represents OD and the last row shows the method OD-CWA.}
	\label{tab:indiv_componets}
\end{table}
We conduct some ablation studies in Table~\ref{tab:indiv_componets} to see how individual components, i.e., SN, CFL, UPL and CWA contribute towards the performance. Adding SN to the method OD improved WI, AOSE by $13.3\% $ and $ 16.4\%$, respectively, and deteriorated the $AP_{\mathcal{U}}$ slightly from OD. Extra analysis of how parameters $\delta$ and $\lambda$ affects the performance of the detector on both ResNet-50 and Swin-T backbone is shown in Table~\ref{tab:ablation_indiv_components} and~\ref{tab:object-detection-performance-swinTr50}. Results pertaining to blur parameter $\nu$ that governs the smoothing/regularisation of Sinkhorn divergences (as per eq~\ref{eq:sinkhorn})  can be found in Table-\ref{tab:Wass_blur}. We select $\nu=1$ based purely on empirical grounds, aiming for optimal performance on all metrics.
\begin{table}[htbp]
 \small
 \centering
     \setlength{\tabcolsep}{2.5pt}
    \renewcommand{\arraystretch}{1.0}
 \begin{tabular}{@{}c|c|@{}c|c|@{}c|c|@{}c|c@{}l|c@{}@{}c|c}
		\hline
SN	& CFL	& CWA & UPL &   $WI_{\downarrow}$ & $AOSE_{\downarrow}$ & $mAP_{K\uparrow}$ & $AP_{U\uparrow}$ \\
		\hline
- & \checkmark	& - & - &  17.92 & 15162 & \textcolor{orange}{58.54} & - \\
-& - 	& - &  \checkmark & 16.47 & 12018 & 57.91 & 14.27\\
	& \checkmark  &  & \checkmark & 14.95 & 11286 & \textbf{58.75} & 14.93 \\
\hline
-	& \checkmark & \checkmark & \checkmark & \textcolor{orange}{12.42}& 10193 & 57.60 & \textcolor{orange}{15.09} \\
\checkmark & \checkmark	&  & \checkmark & 12.96 & \textcolor{orange}{9432}  & {57.86} & 14.78 \\
\checkmark &  \checkmark 	& \checkmark  & \checkmark  & \textbf{11.70}   & \textbf{8748} & {\textbf{57.58}}  & \textbf{15.36} \\
\hline
	\end{tabular}
\caption{Ablation study of individual components for VOC-COCO-$20$ open-set data obtained for R-50 backbone. The third row shows method OD and the last row shows OD-CWA.}
	\label{tab:ablation_indiv_components}
\end{table}
\begin{table}[htbp]
	\small
 \centering
	\setlength{\tabcolsep}{2.5pt}
	\renewcommand{\arraystretch}{1.1}
	\begin{tabular} {c|c|c|c|c|c|c|c|c}
		\hline
	CFL	& CWA & UPL &  ${\frac{\Sigma}{\mu}}_{\downarrow}$ & DI$_{\uparrow}$  & CHI$_{\uparrow}$  & HI$_{\uparrow}$  & DBI$_{\downarrow}$  & XBI$_\downarrow$  \\
\hline
\checkmark	& - & \checkmark &  0.824 & 0.025 & 449.3 & 0.91 & 6.52 & 0.59 \\
\hline
\checkmark 	& \checkmark  & \checkmark  &   {0.04} &  { {0.01}}& {629.27} & { {0.89}} &  {4.24} &  {0.05} \\
\hline
	\end{tabular}
\caption{Quantitative analyses of intra-cluster and inter-clusters variances and distances calculated using different metrics  corresponding to CFL, UPL and CWA on closed-set, i.e., VOC-20 classes for model trained on R-50 backbone.}
	\label{tab:inter&intra}
\end{table}
\begin{table}[htbp]
    \small
    \centering
    \setlength{\tabcolsep}{3.0pt}
    \renewcommand{\arraystretch}{1.05}
    \begin{tabular}{@{}c|c|@{}cc@{}cc@{}cc@{}cc@{}}
        \hline
        $\nu$ & VOC  & \multicolumn{4}{c}{VOC-COCO-20}  \\
        \cline{2-6} 
        & $mAP_{K\uparrow}$ & $WI_{\downarrow}$ & $AOSE_{\downarrow}$ & $mAP_{K\uparrow}$ & $AP_{U\uparrow}$ \\
        \hline
        0.5 & \textbf{79.26} & {12.29} &  {9530} & {57.07} &  {13.39} \\
        \hline
        0.3 & {79.11} & \textcolor{orange}{11.78} & \textcolor{orange}{9175} & {57.31} & {14.51} \\
        \hline
                0.2 &  {78.83}  & {12.17} &  {9178} & {57.44} & \textbf{16.03} \\
                \hline
        0.15 & {79.23}  & {12.23} & {9361} &  \textcolor{orange}{57.46} &  {14.54} \\
        \hline
                \textbf{0.1} &  \textcolor{orange}{79.20}  &  \textbf{11.70} & \textbf{8748} &  \textbf{57.58} &  \textcolor{orange}{15.36} \\
                \hline
    \end{tabular}
    \caption{Effects of variation of blur $\nu$ in the combined loss function for model trained using ResNet-50 backbone and evaluated on VOC-COCO-20 dataset. The blur is a regularisation parameter that governs evaluation of Sinkhorn divergence~\cite{Knight}.}
    \label{tab:Wass_blur}
\end{table}
We conduct some ablation studies to throw light on how each part of the combined loss function Eq.~\eqref{full_equation} behaves, taking into account the compactness of the clusters and how separated are the classes (closed-set) from each other.  The results\footnote{{The intra-class and inter-class evaluation results presented here diverge from those in the study by Han et al.~\cite{han2022opendet}. This disparity stems from the absence of codes necessary to reproduce their results.}} can be seen from the Table~\ref{tab:inter&intra}. {Intra-class variances ($\Sigma$)} are the variances of Euclidean distances obtained between the proposal embeddings of detections and the centroid of a particular class. {Inter-class distance ($\mu$)} average of Euclidean distance between all the class centres of different classes. The ratio of $\frac{\Sigma}{\mu} = \frac{\text{Var} (\mathbb{E}_j(\mathbf{z}_{j,d} - \mathbb{E}_d(\mathbf{z}_{j,d}) ) )}{  \sum_{j,k \in \{1,C\}, j \neq k }  {\left\lVert \mathbf{z}_i - \mathbf{z}_j \right\rVert}_2}$ is reported in Table-\ref{tab:inter&intra}. Along with this ratio, we also add a couple of other widely used indices, e.g., Dunn Index (DI), Calinski-Harabasz Index (CHI), Hubert Index (HI), Davies Bouldin index (DBI) and Xie-Beni Index (XBI). The details of these indices can be found in~\cite{Desgraupes2016ClusteringI}. It is important to recognise that there isn't a single optimal metric for evaluating intra-cluster and inter-cluster properties, so we have chosen a few popular ones. When, CWA is added to CFL and UPL, there is a slight drop in performance for two out of the six indices, and good improvement in the rest indices. 
\section{Conclusion} \label{sec:conclusion}
A novel open-set detector, OD-CWA specifically designed to address the challenge of open-set detection is presented. OD-CWA aims to enhance the discrimination of low-density latent regions and improve clustering through the integration of three key components: IC loss, CWA loss, and UP loss. Both CWA and \textcolor{black}{IC} components are guided by spectral normalisation of weights in the final output layer. These two losses along with UP loss further enhances the model's capability in flagging unknown objects in the test instances. The promising empirical results and a couple of novel additions suggest that this approach has the potential to pave the way for new research directions in OSOD.

\textbf{Limitations:} We notice that this approach still suffers from reasonable open-set errors, especially when there are a lot of objects in the image. Sometimes, there are instances where the detector is confused during inference on an unknown class image and its predictions are both unknown with high score and known with medium prediction score. We aim to redirect our attention towards rediscovering $\mathcal{L}_{UP}$ in light of this observed confusion. Furthermore, as Wasserstein distance approach is new and comes with some promising theoretical results, our future work will predominantly centre on harnessing these guarantees to derive generalisation bounds for open-set conditions. 

{
    \small
    \bibliographystyle{ieeenat_fullname}
    \bibliography{main}
}

\clearpage
\setcounter{page}{1}
\counterwithin{figure}{section}
\counterwithin{table}{section}
\renewcommand\thefigure{\thesection\arabic{figure}}
\renewcommand\thetable{\thesection\arabic{table}}
\appendix
\maketitlesupplementary
\section{Visualisation of proposal embeddings}
\begin{figure}[htbp]
    \centering
   { \subfloat[\centering OD]{{\includegraphics[width=3.7cm]{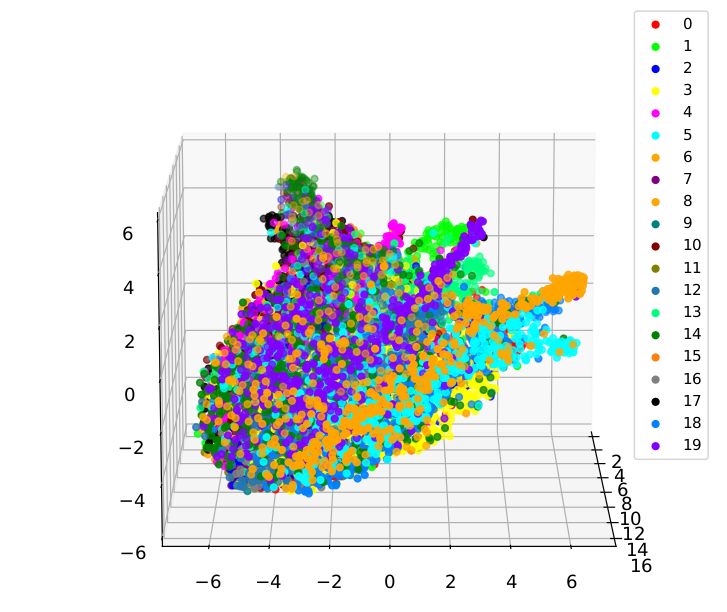}}}}
    \qquad
   {\subfloat[\centering OD-CWA]{{\includegraphics[width=3.7cm]{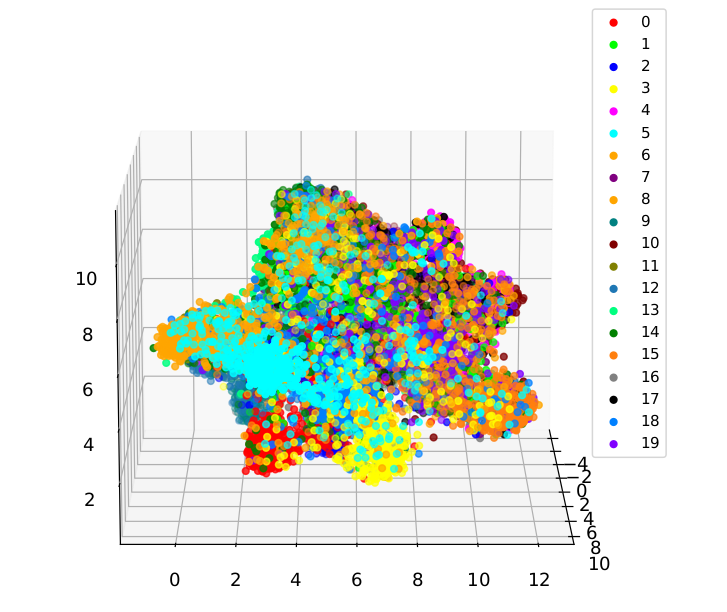} }}}
      { \subfloat[\centering OD]{{\includegraphics[width=3.7cm]{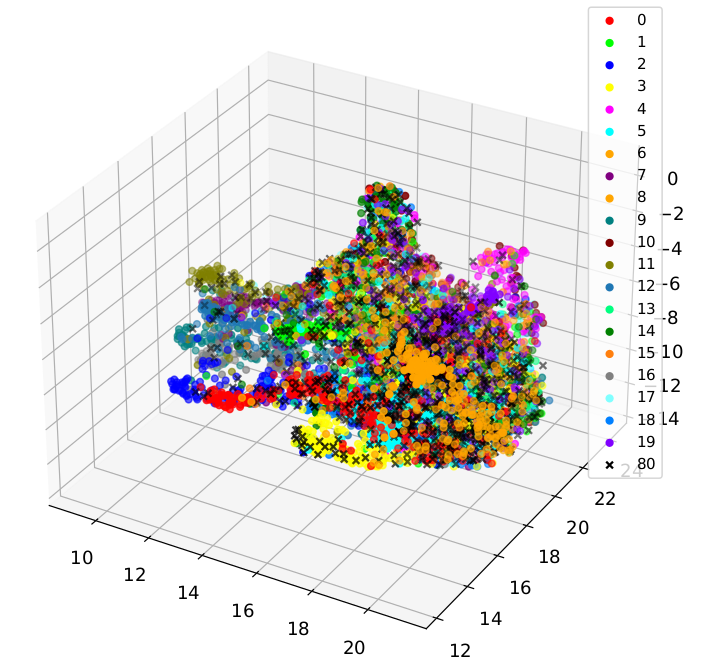}}}}
    \qquad
   {\subfloat[\centering OD-CWA]{{\includegraphics[width=3.7cm]{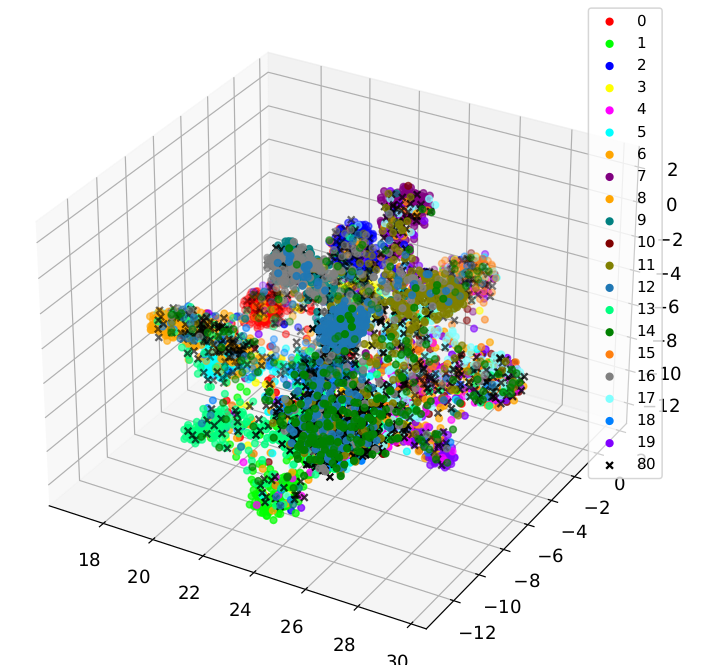} }}}%
    \caption{A full U-MAP~\cite{mcinnes2020umap} visualisation of proposal embeddings $\mathbf{F}_d \in \mathbb{R}^{128}$, $d$ refers to number of detections, for VOC-20 closed-set classes. Top two subplots shows all detections of known objects and bottom two subplots shows subset of known along with unknown embeddings to show scattering of low-density regions.}%
    \label{fig:logits_full}%
\end{figure}

Figure-\ref{fig:logits_full} provides a U-MAP~\cite{mcinnes2020umap}-based visual representation of proposal embeddings for closed-set ($C_{\mathcal{K}}$) class detections. The top two subplots demonstrate the reduced-dimensional proposal embeddings of detections for baseline (OD) and OD-CWA approaches. OD-CWA exhibits more cohesive clusters compared to OD. In the bottom two subplots, we narrow our focus to a subset of detections from the top plots and overlay unknown detections (depicted by black $\pmb{\times}$ symbols) to emphasise the dispersion of unknowns amid known objects. This shows the scattering of low-density latent regions. The plots are appropriately rotated for improved visibility.
\section{Additional Experimental Information}
\begin{figure}[htbp]
    \centering
     {\includegraphics[width=6.6cm]{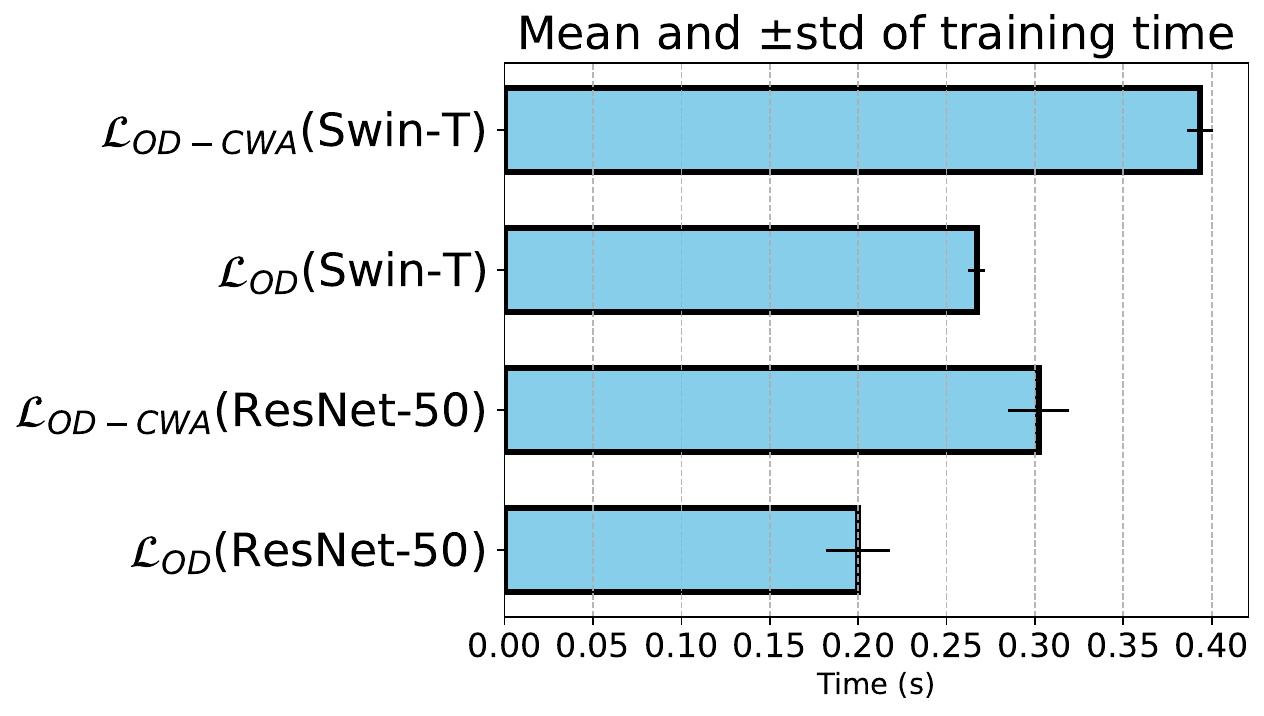}}
    \caption{Training time for models trained with loss functions, $\mathcal{L}_{OD}$ and $\mathcal{L}_{OD-CWA}$ taking into account two different backbones, i.e., ResNet-50 and Swin-T. The training time stamps are collected in seconds and measured per 20 iterations ran on 8 gpus.}%
    \label{fig:trianing_time}%
\end{figure}

This section provides some comprehensive results, i.e.,

\begin{enumerate}
\item elaboration on additional experimental details that were omitted in the main paper due to space limitations. This includes the presentation of two tables (Table~\ref{tab:full-swinTr50} and~\ref{tab:full-swinTr50_coco_voc0.5n}) related to performance of Swin-T backbone (part of which was in the main paper). Table~\ref{tab:suppl_table_comp_resnet_swin} summarises performance of OD and OD-CWA based on two different backbones on VOCO-COCO-20.

\item training time aspects of two distinct approaches, namely OD and OD-CWA can be found in Fig-\ref{fig:trianing_time}. Each approach is examined under two different backbones. 
\item Inclusion of pertinent details for reproducing the codes developed by Han et al.~\cite{han2022opendet}. This section serves as a valuable resource for researchers seeking to replicate and validate the outcomes of our study, fostering transparency and reproducibility in scientific endeavours.
\end{enumerate}
\begin{table*}[ht]
\small
     \setlength{\tabcolsep}{1.8pt}
    \renewcommand{\arraystretch}{1.02}
    \centering
    \begin{tabular} {@{}|c|@{}@{}c|c|@{}@{}cc@{}@{}cc|c@{}@{}c@{}c@{}@{}c|c@{}@{}c@{}c@{}@{}c|}
        \hline
          $\delta_k$ & $\lambda$ & VOC  & \multicolumn{4}{c|}{VOC-COCO-20}  & \multicolumn{4}{c|}{VOC-COCO-40} & \multicolumn{4}{c|}{VOC-COCO-60} \\
        \cline{3-15}    
        & & $mAP_{K\uparrow}$ & $WI_{\downarrow}$ & $AOSE_{\downarrow}$ & $mAP_{K\uparrow}$ & $AP_{U\uparrow}$ & $WI_{\downarrow}$ & $AOSE_{\downarrow}$ & $mAP_{K\uparrow}$ & $AP_{U\uparrow}$ & $WI_{\downarrow}$ & $AOSE_{\downarrow}$ & $mAP_{K\uparrow}$ & $AP_{U\uparrow}$ \\
        \hline
          0.25 &  1.3 &  \textbf{83.61}         & {10.57}      & {9038}     & {63.46 }      & { 13.86 }  & 12.78  & 12880  &  61.18 & 9.96 &  9.73  & 13079 & \textcolor{orange}{62.08} &  3.47  \\
             0.18  & 1.3 & {83.06}         & \textbf{10.32}      & \textcolor{orange}{8888}     & {62.93}      & {15.34}   & 12.91 & \textcolor{orange}{12700} &   60.57  & 10.73 &  9.91 & 13226 & 61.58  & \textcolor{orange}{3.82}  \\
         0.21 & 1.6 &   {79.20} & {12.16} & {10007} & {57.09} & \textcolor{orange}{15.53} & \textbf{12.44} & 12792  &  \textcolor{orange}{61.24} &   \textcolor{orange}{10.80} & \textbf{9.61} & 13178 &  62.02 &  3.79 \\
             0.21  & 1.83 & {83.15}         & {10.44}      & {8951}     & \textcolor{orange}{63.20}      & {15.41}  & \textcolor{orange}{12.55} & \textbf{12667}  &   60.70 & 10.59 & \textcolor{orange}{9.71} & \textcolor{orange}{13065} & 61.60 &  3.79  \\
             \hline
        {0.21}  & {1.7} & \textcolor{orange}{83.34}         & \textcolor{orange}{10.35}      & \textbf{8946}     & \textbf{63.59}      & \textbf{18.22}  & 12.73  & 12710 &  \textbf{61.26} &  \textbf{12.02} &  9.76  & \textbf{12998} & \textbf{62.23} &  \textbf{4.28}  \\
        \hline
    \end{tabular}
    \caption{Additional results of Swin-T backbone-based proposed method OD-CWA evaluated on VOC and VOC-COCO-$\{\textbf{T}_1\}$.  This was skipped in the main paper because of space constraints, however, the first $4$ columns were shown in the main paper.}
    \label{tab:full-swinTr50}
\end{table*}
\begin{table*}[htbp]
\small
     \setlength{\tabcolsep}{1.8pt}
    \renewcommand{\arraystretch}{1.02}
    \centering
    \begin{tabular} {@{}|c|@{}@{}c|@{}@{}cc@{}@{}cc|c@{}@{}c@{}c@{}@{}c|c@{}@{}c@{}c@{}@{}c|c@{}@{}c@{}c@{}@{}c|}
        \hline
          $\delta_k$ & $\lambda$ &  \multicolumn{4}{c|}{VOC-COCO-$0.5n$}  & \multicolumn{4}{c|}{VOC-COCO-$n$} & \multicolumn{4}{c|}{VOC-COCO-$2n$} & \multicolumn{4}{c|}{VOC-COCO-$4n$} \\
        \cline{3-18}    
        &   & $WI_{\downarrow}$ & $AOSE_{\downarrow}$ & $mAP_{K\uparrow}$ & $AP_{U\uparrow}$ & $WI_{\downarrow}$ & $AOSE_{\downarrow}$ & $mAP_{K\uparrow}$ & $AP_{U\uparrow}$ & $WI_{\downarrow}$ & $AOSE_{\downarrow}$ & $mAP_{K\uparrow}$ & $AP_{U\uparrow}$ & 
        $WI_{\downarrow}$ & $AOSE_{\downarrow}$ & $mAP_{K\uparrow}$ & $AP_{U\uparrow}$ \\
        \hline
          0.25 &  1.3          & {3.98}      & {2425}     & {83.56}      & { 6.93 }  &7.96   & 5303  &  \textbf{80.62} & 10.23 &14.22 & 10666 & \textbf{76.28} &   13.17  & 21.80 & 21390 &  \textcolor{orange}{70.27} & 15.37 \\
             0.18  & 1.3        & \textcolor{orange}{3.98}      & \textcolor{orange}{2538}     & {82.90 }      & {7.70 }   & 8.02 & 5252 &  79.89 & 11.11  & 14.19 & 10672 & 75.61  & \textcolor{orange}{13.85} & 22.07 & 21451 & 69.59 &  15.96   \\
         0.21 & 1.6   & {4.02} & {2427} & {83.33 } &  7.59 &  \textcolor{orange}{7.82} &  5210 & 80.25  & 10.75 & \textcolor{orange}{13.91}   & 10673 & 76.00  &  13.75 & \textcolor{orange}{21.68} & 21286 &  70.07 & \textcolor{orange}{16.02} \\
             0.21  & 1.83          & 4.26    & 2377     & \textcolor{orange}{82.91}      & \textcolor{orange}{8.11 }  &  8.21& \textbf{5079}  &   79.91 & \textcolor{orange}{11.22} & 14.44 & \textbf{10464} & 75.73 &  \textcolor{orange}{13.85} &  22.55 & \textbf{20764} & 70.16  & 15.81 \\
             \hline
        {0.21}  & {1.7}           & \textbf{3.73}      & \textbf{2429}     & \textbf{83.24 }      & \textbf{ 9.45 }  & \textbf{7.36}   & \textcolor{orange}{5162} &  \textcolor{orange}{80.25}  &  \textbf{12.67} &  \textbf{13.14}  & \textcolor{orange}{10576} &  \textcolor{orange}{76.08} &  \textbf{15.01} & \textbf{20.64} & \textcolor{orange}{21255} &   \textbf{70.55} & \textbf{16.87} \\
        \hline
    \end{tabular}
    \caption{Additional results of Swin-T backbone-based proposed method OD-CWA evaluated on VOC and VOC-COCO-$\textbf{T}_2$ i.e., disjoint sets $\{0.5n, n, 2n, 4n\}$. These results were skipped in the main paper because of space constraints. VOC-COCO-$2n$ is not a separate dataset, rather a part of VOC-COCO-$\{\textbf{T}_2\}$~\cite{han2022opendet}.}
    \label{tab:full-swinTr50_coco_voc0.5n}
\end{table*}
\begin{figure}[htbp]
     {\includegraphics[width=8cm]{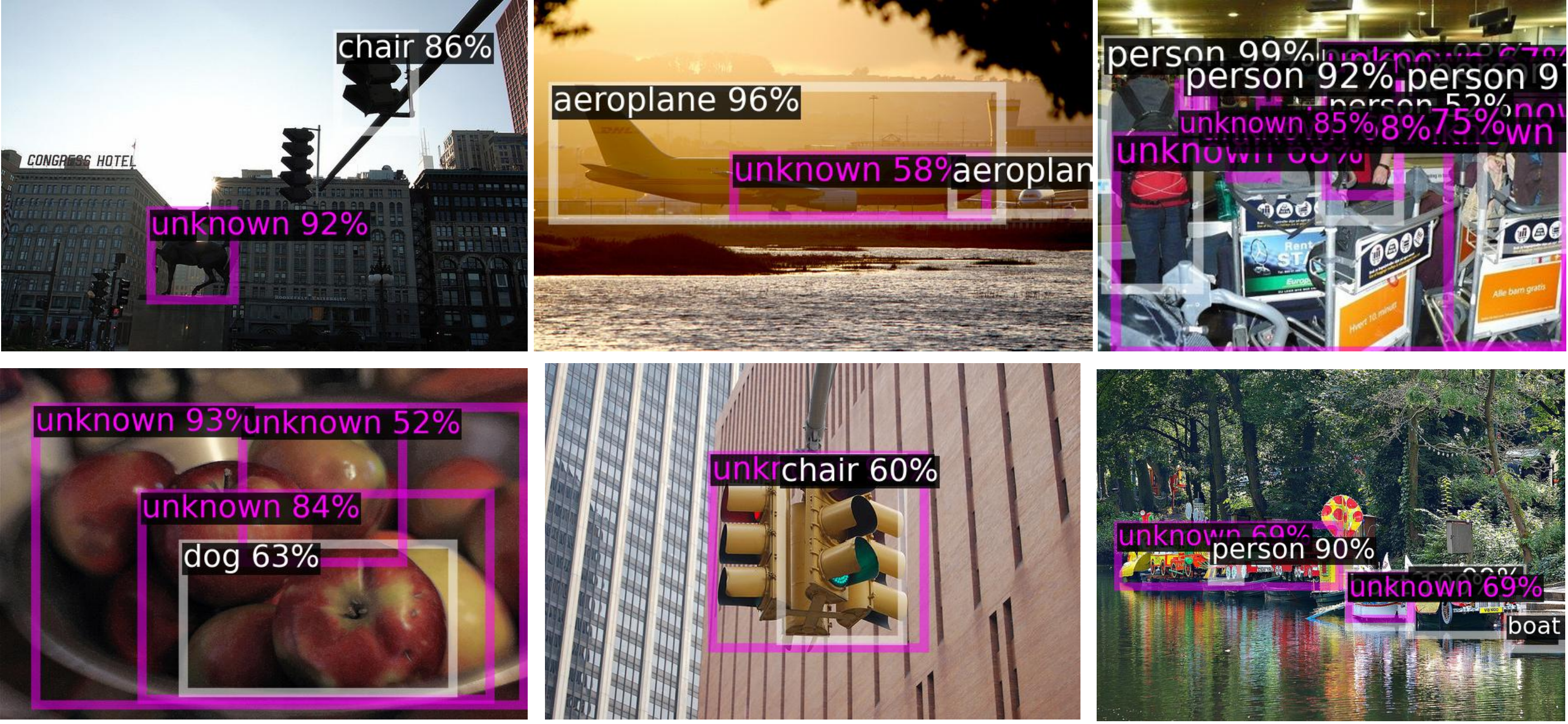}}
    \caption{Failure Cases for OD-CWA. It does not imply the success by OD, rather it means there were misclassifications and confusion arising from cluttered images.}%
    \label{fig:failure}%
\end{figure}
\subsection{Experimental Setup Details}
The experimental setup used for training and evaluation is outlined in this section. To optimise the utilisation of high-performance computing resources, we conducted our experiments on \texttt{NVIDIA DGX} clusters integrated with Kubernetes. The \texttt{NVIDIA DGX} systems, featuring \texttt{Tesla V100-SXM2-32}GB GPUs, formed our computing infrastructure. The combination of PyTorch, Kubernetes, and NVIDIA DGX contributed to the efficiency of our experiments, enabling expedited training and evaluation of models within a distributed computing environment. 
\subsection{Comments on utilising codes by Han~\etal~\cite{han2022opendet}}
{All qualitative results presented herein, as well as those detailed in the main paper, were derived through the training of a model utilising the $\mathcal{L}_{\text{OD-CWA}}$ for $80,000$ iterations. The inference phase, performed to facilitate the comparison of results on a set of images, is depicted in Fig-\ref{fig:inference_images_new}. The model training employed precise parameters consistent with the OD framework~\cite{han2022opendet}. However, attempting to execute OD by replicating the provided codes from the GitHub repository authored by Han et al.~\cite{han2022opendet} revealed issues, necessitating modifications for successful run. Challenges such as version mismatches resulted in extended errors. To address this in future, we will be releasing codes (upon acceptance) which leverages docker for model training, ensuring a seamless and error-free execution environment that alleviates version discrepancies, guaranteeing a consistent experimental setup.}
\subsection{Comprehensive Qualitative Comparison}
\begin{figure*}[htbp]
    \centering
     {\includegraphics[width=15cm]{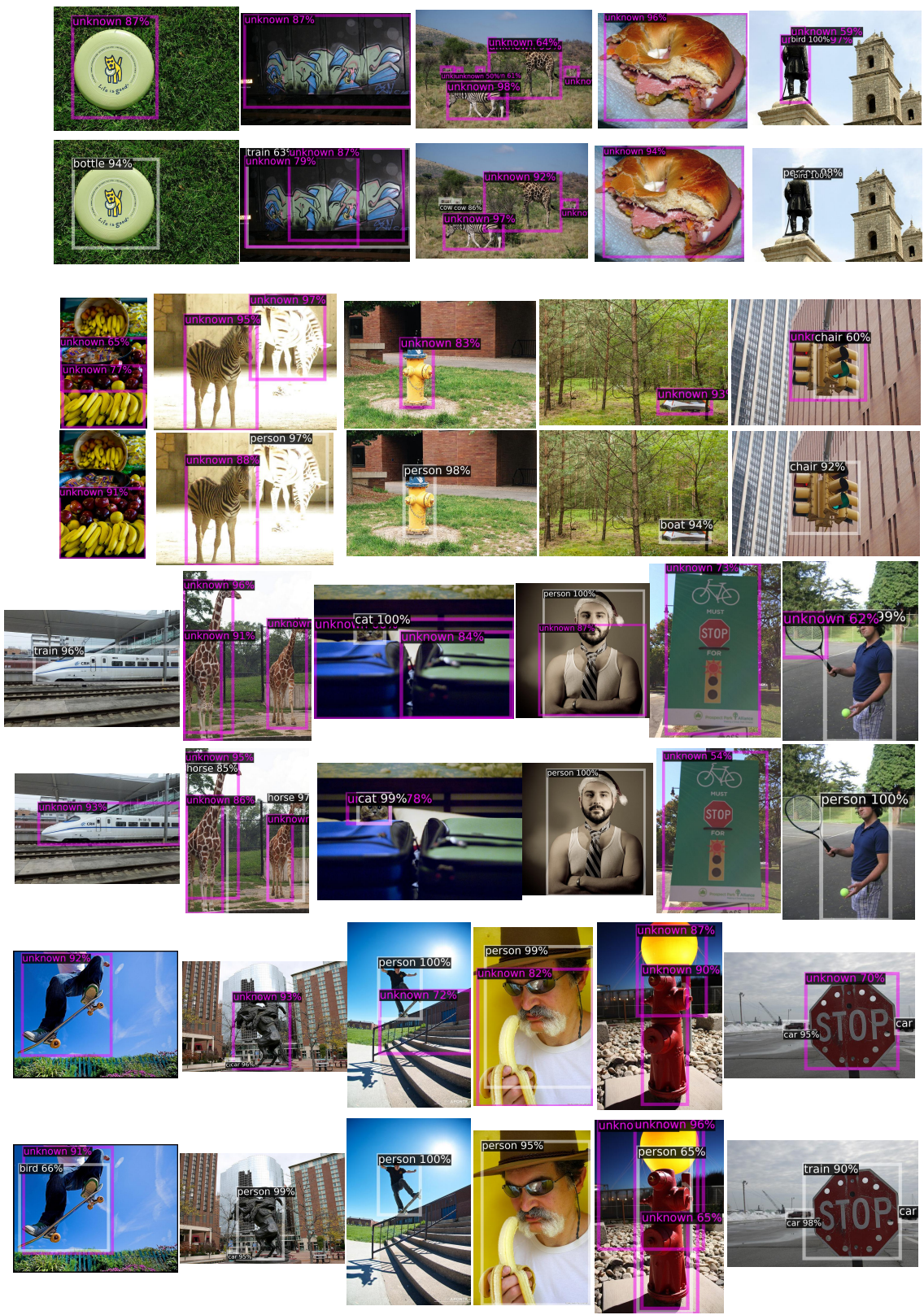}}
    \caption{Qualitative comparisons between OD and proposed OD-CWA. In each pair of images, the top image represents OD-CWA, while the other corresponds to OD. Objects marked in purple indicate unknown entities, while white denotes known objects. For instance, in the first image pair, the object labelled as a Frisbee (unknown) is misclassified as a bottle by the model trained with OD~\cite{han2022opendet}. In contrast, OD-CWA correctly identifies the object as belonging to an unknown class.}%
    \label{fig:inference_images_new}%
\end{figure*}

{Fig-\ref{fig:inference_images_new} provides a detailed qualitative analysis comparing the baseline method OD with our developed approach OD-CWA. OD excels in detecting unknowns, evident in figure pairs $2, 3, 4, 5, 6, 7, 12, 13, 15, 17, 21$. However, OD-CWA surpasses these capabilities by not only enhancing the prediction scores of unknowns but also reducing confusion. For instance, in figure pair 7, OD detects one zebra as a known object (person) and the other as unknown with a predictive score of $88\%$. In contrast, OD-CWA confidently predicts both zebras as unknown with much higher predictive power. Similar improvements are observed in other instances, such as figure pairs $12, 13, 15$, and more.}

\begin{table}[htbp]
    \small
    \centering
    \setlength{\tabcolsep}{1.4pt}
    \renewcommand{\arraystretch}{1}
    \begin{tabular}{c|c|c|c|c|c} 
        \toprule
        Method & Backbone & $WI_{\downarrow}$ & $AOSE_{\downarrow}$ & $mAP_{K\uparrow}$ & $AP_{U\uparrow}$ \\
        \midrule
        \multirow{2}{*}{OD} & ResNet-50  & 14.95 &  11286 & \textbf{58.75} & 14.93 \\
                                       & Swin-T  &   12.51 &  9875  & 63.17 & 15.77 \\
        \hline
       \multirow{2}{*}{OD-CWA} & ResNet-50 & \textbf{11.70} & \textbf{8748} & 57.58 & \textbf{15.36}\\
        & Swin-T  & \textbf{10.35} & \textbf{8946} & \textbf{63.59} & \textbf{18.22} \\
        \bottomrule
    \end{tabular}
    \caption{Summary of OD and OD-CWA results on ResNet-$50$ and Swin-T backbones. Bold represents the best of two methods}
    \label{tab:suppl_table_comp_resnet_swin}
\end{table}
\subsection{Failure Cases}
Fig-\ref{fig:failure} showcases instances considered as failures when tested by our method OD-CWA. Six image examples are provided. In the first image, OD-CWA identifies the statue as unknown but fails to recognise distinctive objects such as traffic lights. In other cases, such as the $2^{nd}$ image, the detector identifies an aeroplane but is confused by the flight engines, mistaking them for an unknown object. Similarly, in the $4^{th}$ instance, the detector categorises traffic lights as unknown, but a part of the lights is identified as a chair. It's important to note that even though these instances are considered failures, it doesn't imply correct identification by the baseline (OD); in fact, the baseline inference results were worse for these cases.

\end{document}